\documentclass[letterpaper]{article} 
\usepackage{aaai25}  
\usepackage{times}  
\usepackage{helvet}  
\usepackage{courier}  
\usepackage[hyphens]{url}  
\usepackage{graphicx} 
\usepackage{natbib}  
\usepackage{caption} 
\usepackage{algorithm}
\usepackage{algorithmic}
\usepackage{dsfont}
\usepackage{amsmath,amsfonts}
\usepackage{algorithmic}
\usepackage{algorithm}
\usepackage{array}
\usepackage{subcaption}
\usepackage[caption=false,font=normalsize,labelfont=sf,textfont=sf]{subfig}
\usepackage{textcomp}
\usepackage{amssymb}
\usepackage{url}
\usepackage{verbatim}
\usepackage{graphicx}
\usepackage{xcolor}
\usepackage{cite}
\usepackage{multirow}
\usepackage{colortbl}
\usepackage{arydshln}

\usepackage{newfloat}
\usepackage{listings}
\DeclareCaptionStyle{ruled}{labelfont=normalfont,labelsep=colon,strut=off} 
\floatstyle{ruled}
\newfloat{listing}{tb}{lst}{}
\floatname{listing}{Listing}
\title{Rethinking Pseudo-Label Guided Learning for Weakly Supervised Temporal Action Localization from the Perspective of Noise Correction}
\author {
    Quan Zhang\textsuperscript{\rm 1}\equalcontrib,
    Yuxin Qi\textsuperscript{\rm 2,3}\equalcontrib,
    Xi Tang\textsuperscript{\rm 1},
    Rui Yuan\textsuperscript{\rm 1},
    Xi Lin\textsuperscript{\rm 2,3},
    Ke Zhang\textsuperscript{\rm 4}\thanks{Corresponding author.},
    Chun Yuan\textsuperscript{\rm 1}\footnotemark[2]
}

\affiliations {
    \textsuperscript{\rm 1}Tsinghua Shenzhen International Graduate School, Tsinghua University, Shenzhen, China\\
    \textsuperscript{\rm 2}School of Electronic Information and Electrical Engineering, Shanghai Jiao Tong University, Shanghai, China\\
    \textsuperscript{\rm 3}Shanghai Key Laboratory of Integrated Administration Technologies for Information Security, Shanghai, China\\
    \textsuperscript{\rm 4}School of Computer Science and Technology, Soochow University, Suzhou, China\\
    \{zhangqua22, tangx24, yuanr22\}@mails.tsinghua.edu.cn, \{qiyuxin98,linxi234\}@sjtu.edu.cn, 
kzhang19@suda.edu.cn, yuanc@sz.tsinghua.edu.cn

}

\usepackage{bibentry}

\begin{document}

\maketitle

\begin{abstract}
Pseudo-label learning methods have been widely applied in weakly-supervised temporal action localization. Existing works directly utilize weakly-supervised base model to generate instance-level pseudo-labels for training the fully-supervised detection head. We argue that the noise in pseudo-labels would interfere with the learning of fully-supervised detection head, leading to significant performance leakage. Issues with noisy labels include:(1) inaccurate boundary localization; (2) undetected short action clips; (3) multiple adjacent segments incorrectly detected as one segment. To target these issues, we introduce a two-stage noisy label learning strategy to harness every potential useful signal in noisy labels. First, we propose a frame-level pseudo-label generation model with a context-aware denoising algorithm to refine the boundaries. Second, we introduce an online-revised teacher-student framework with a missing instance compensation module and an ambiguous instance correction module to solve the short-action-missing and many-to-one problems. Besides, we apply a high-quality pseudo-label mining loss in our online-revised teacher-student framework to add different weights to the noisy labels to train more effectively. Our model outperforms the previous state-of-the-art method in detection accuracy and inference speed greatly upon the THUMOS14 and ActivityNet v1.2 benchmarks.
\end{abstract}

\section{Introduction}

Temporal  action localization (TAL) \cite{chao2018rethinking} is a significant  task in video comprehension, which aims to locate the start and end times of target action instances within untrimmed videos and identify their categories. It has shown great potential in tasks like intelligent surveillance, video summarization, etc. 
Inspired by the widespread application of deep learning methods across various fields\cite{chen2024gim,zhang2024distilling,lu2024cricavpr,lu2024deep,lu2024towards}, temporal action localization has widely adopted deep learning-based approaches, which has sparked significant attention.
Although the TAL problem has garnered much attention, current prominent methods often rely on costly, labor-intensive temporal boundary annotations, hindering the widespread applications of TAL.

To reduce reliance on instance-level labels, Weakly Supervised Temporal Action Localization (WTAL) solely based on video-level labels \cite{chen2022dual, zhang2024can} has gained increasing attention. WTAL only requires knowledge of which actions occurred in the video, without the need for precise location information.

\begin{figure}[t]
	\centering
	\includegraphics[width=1.0\linewidth,height=6cm]{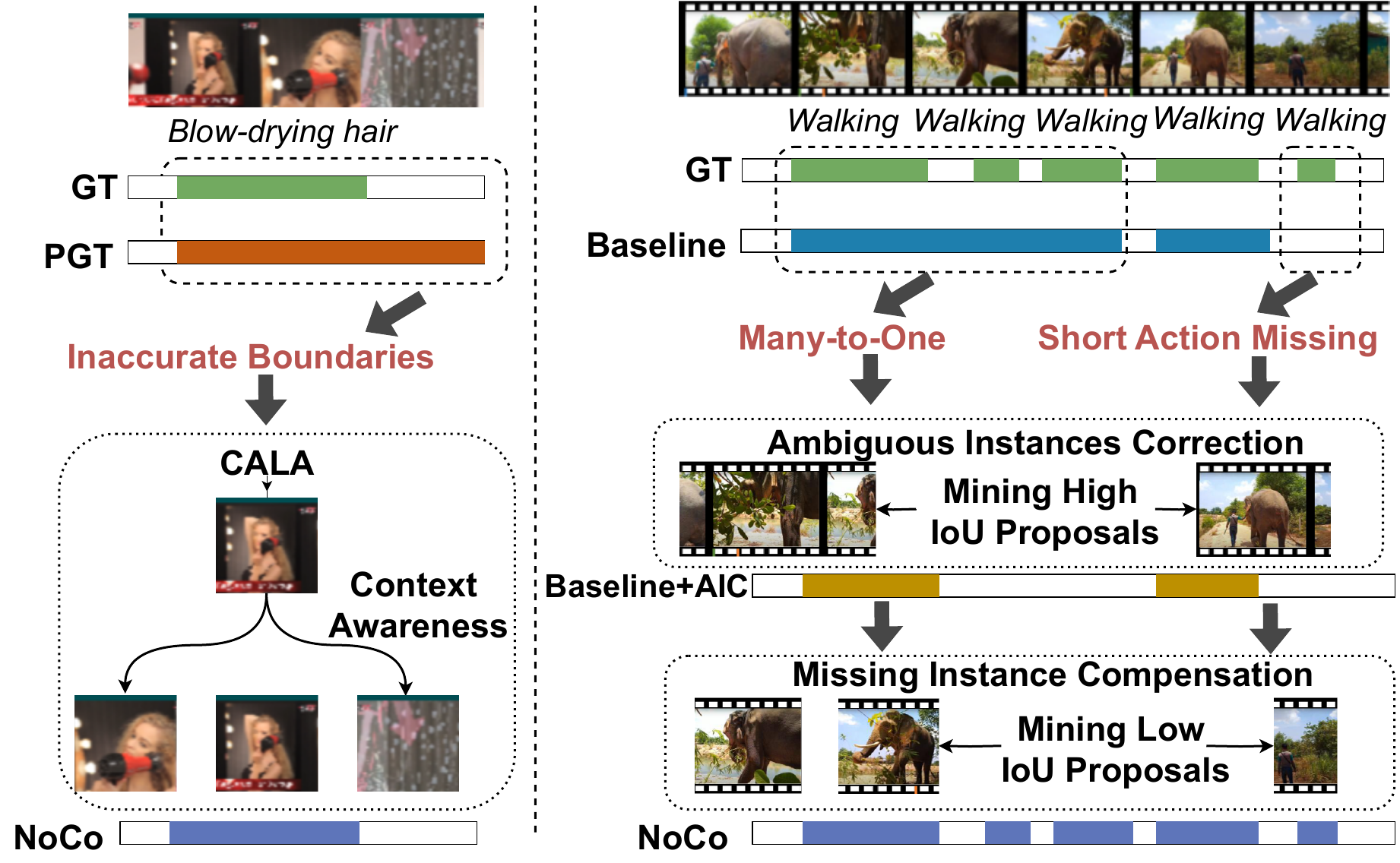}
	\caption{Noise correction modeling for WTAL. NoCo introduces pseudo-label noise correction modules to address the typical failures in existing WTAL methods.}
	\label{intro figure}
\end{figure}

Existing WTAL works mainly follow per-category localization pipelines \cite{wang2017untrimmednets, zhao2022equivalent}, using video-level classifiers trained on class labels \cite{narayan2021d2} to filter each video clip. However, due to the absence of refinement mechanisms, these models attribute high confidence to incorrect segments, such as background elements, leading to imprecise localization. 
Many studies \cite{liu2019completeness, liu2021weakly, liu2021acsnet} attempt to address the disparity between classification and localization, with a promising solution being the generation and utilization of pseudo-labels which benefits from utilizing clip-level supervision instead of video-level labels. Existing works \cite{yang2021uncertainty, luo2020weakly, pardo2021refineloc, zhai2020two} have achieved remarkable results by introducing pseudo-labels on WTAL. 

However, due to the sparsity of video-level label supervision information, there exists a large amount of noises in pseudo-labels. We argue that these noises can be classified into three types: 1) Inaccurate instance boundaries; 2) Missing action instances; 3) Adjacent instances aggregated into one instance, as shown in Figure \ref{intro figure}. \par
To mitigate the impact of noisy pseudo-labels on WTAL, we reevaluate the pseudo-label learning process from the perspective of noise correction. We propose NoCo, a weakly-supervised temporal action localization method, which trains a weakly-supervised pseudo-label generator first, and then incorporate proposed teacher-student-based online noise learning framework over noisy pseudo-labels.
In practice, there exists two issues in the above two stages. (1) How to generate accurate pseudo-label. (2) How to implement the online noise correction framework.\par
To generate accurate pseudo-labels, we argue that using Non-Maximum Suppression (NMS) \cite{neubeck2006efficient} loses the contextual information, leading to inaccurate instance boundaries. In the pseudo-label generation process, contextual information serves as a valuable supplementary source. 
Motivated by this, we introduce the Context-Aware Label Augmentation module (CALA), which adaptively perceives the contextual information for each pseudo-label and and corrects them through weighted averaging. Such augmentation not only yields more accurate action boundaries but also provides a more reasonable initial pseudo-label set for subsequent noise correction framework.\par
To realize online noise correction, we introduce a strategy that combines noise correction with teacher-student training. 
This approach leverages the potential of the teacher-student framework in noise learning, allows for the convenient addition of modules tailored for WTAL, and ensures high inference efficiency.
Different from the previous teacher-student strategy, our approach introduces several key innovations. Firstly, to generate robust and stable pseudo-label compensation information, our teacher model employs weighted aggregation of multiple historical student models.  
Secondly, we propose a Missing Instance Compensation module (MIC) to solve the short clips missing problem by identifying compensation instances with low intersection-over-union (IoU) ratio from the teacher prediction information.
Thirdly, the Ambiguous Instance Correction module (AIC) is introduced to address the issue of aggregating adjacent instances into one by mining the high IoU ratio instance from the teacher's online pseudo-labels. Lastly, a High-quality Pseudo-label Mining loss (HPM) is proposed to construct adaptive optimization weights, fully leveraging high-quality supervision information.

It should be noted that our noise correction framework is decoupled from the WTAL base model. The inference process of NoCo does not depend on the WTAL base model.In this paper, we build a student model based on the efficient TAL model TriDet \cite{shi2023tridet}, and during the inference phase, only the student model needs to be run.
\par
In summary ,the main contributions are listed as follows.
\begin{itemize}
    \item 
    We revisit the WTAL process from the perspective of noise correction and propose NoCo, a noise learning framework aimed at gradually mitigating the pseudo-label noise effect in TAL. We observe three types of noise information present in the pseudo-label learning method.

    \item 
    We propose a Context-aware Label Augmentation module  aimed at rectifying inaccurate action boundaries. We present a teacher-student-based online noise correction framework to address short-action omissions and many-to-one problems, encompassing Missing Instance Compensation, Ambiguous Instance Correction , and High-quality Pseudo-label Mining.

    \item  We conduct extensive experiments to demonstrate the effectiveness of NoCo. Our methods bring a significant improvement for WTAL baselines on two benchmarks and achieve state-of-the-art performance. Additionally, NoCo demonstrates fast inference speed.

\end{itemize}

\section{Related Work}
\noindent\textbf{Weakly-supervised temporal action localization} aims to determine the start and end times, as well as the category, of specified action instances in untrimmed videos using only video-level labels. Unlike traditional fully-supervised methods, it doesn't rely on costly instance-level labels, making it a widely studied approach\cite{paul2018w}.
Existing weakly-supervised methods fall into four categories: a) Multi-instance Learning: UntrimmedNet\cite{wang2017untrimmednets} was first applied to WTAL, generating action proposals which were then classified. However, such methods tend to focus on the most discriminative regions due to the difference between classification and localization tasks. b) Metric Learning: WTALC\cite{paul2018w} utilizes metric learning to make features of the same class similar. c) Attention-Based Learning: STPN\cite{nguyen2018weakly} introduced a sparse loss for attention sequences, capturing key foreground segments. d) Pseudo-label Learning: TSCN\cite{zhai2020two} generates segment-level pseudo-labels by fusing attention from RGB and FLOW modalities,while UGCT \cite{yang2021uncertainty} considers their complementarity and ASM-Loc \cite{he2022asm} enhances features with action proposals. Previous pseudo-label learning methods have made great progress, but did not consider the existence of noise in pseudo-labels.We are the first to improve pseudo-label learning methods from the perspective of noise correction.
\par

\begin{figure*}[h]
	\centering
	\includegraphics[width=17.5cm,height=8.2cm]{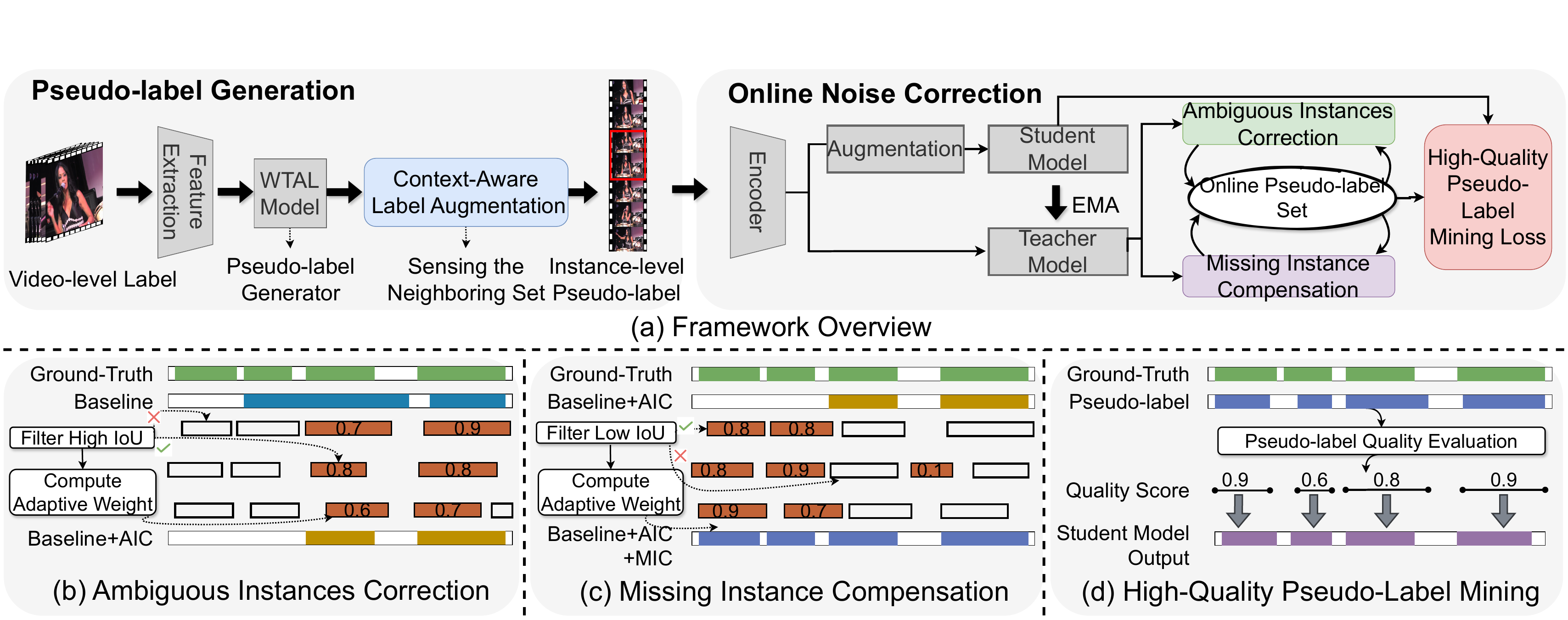}
	\caption{(a). Framework Overview. The dark gray modules indicate the base method (eg. WTAL model and augmentation), while the others are our noise correction modules. (b). Ambiguous Instances Correction is based on teacher predictions to accurate instance boundaries through mining high IoU samples and adaptively aggregating context. It is able to solve many-to-one and in-accurate boundary position problem. (c). Missing Instance Compensation focuses on adding missing instance based on teacher predictions. It aims to solve short action missing problem. (d). High-Quality Pseudo-label Mining Loss assigns adaptive weights for each noisy action instance to mine high-quality pseudo-label.}
	\label{pipeline}
\end{figure*}

\noindent\textbf{Pseudo Label Guided Training} has been widely applied in visual tasks with weak or limited supervision. In weakly-supervised object detection, one common approach is self-training \cite{ren2020instance}. This method involves training an teacher model first and using high-confidence predictions to generate instance-level pseudo-labels. These pseudo-labels are then utilized to train the final detector.
Similarly, in semi-supervised learning \cite{weng2022semi} and domain adaptation \cite{das2018graph}, models are first trained on labeled/source datasets and then used to generate pseudo-labels for unlabeled/target datasets, guiding the training process. Similar to these works, our method utilizes segment-level pseudo-labels to guide  training process for the WTAL task. The difference is that we notice noises in the pseudo-label, and progressively correct these noises.
\par
\noindent\textbf{Learning with Noisy Labels} has been widely applied in image classification tasks. To effectively leverage noise labels to guide the training process, DivideMix \cite{li2020dividemix} uses two networks for sample selection via binary mixture modeling. Price et al. \cite{pleiss2020identifying} introduced the marginal area statistic, measuring the average difference between a sample's assigned class logit and its highest unspecified class logit, to separate correctly from incorrectly labeled data. Liu et al. \cite{liu2020early} found the model initially predicts ground-truth labels but eventually memorizes wrong ones, motivating exploiting the model's early output. Motivated by these works, we propose noisy label learning framework for TAL task.

\section{Methods}
In this section, we introduce the proposed NoCo framework in detail. First, we present the problem definition and describe the framework overview. Then we present the main components of NoCo in detail. \par
\textbf{Problem Definition.}
Given an untrimmed training video $V_i$ and its video-level label $y_i \in \{0, 1\}^C$, where $y_i$ is an C-dimension one-hot vector indicating category. The goal of the inference phase is to predict an action instance set $\mathcal{A} = \{(c_i,\theta_i,s_i,e_i)\}$, where $c_i$ denotes the category of the instance $A_i \in \mathcal{A}$, $\theta_i$ is the confidence score, and $s_i$ and $e_i$ denote the start and end timestamp.\par

\textbf{Framework Overview.} As illustrated in Figure \ref{pipeline}, our noise correction framework comprises two main stages. Firstly, the WTAL Base Model generates noisy pseudo-labels and then CALA is employed for context-aware enhancement of noisy pseudo-labels to achieve more accurate boundaries. Secondly, the enhanced pseudo-labels are input into the online noise correction framework as the initial pseudo-label set.\par

In the online correction framework, the teacher model is aggregated from various historical versions of student models, providing stable compensatory information for noise correction. The Missing Instance Compensation module (MIC) identifies action instances with low IoU with the pseudo-label, generating compensatory action instances. Simultaneously, the Ambiguous Instance Correction module (AIC) discovers action instances with high IoU with the pseudo-label, creating an associated action instance set. Then this associated set is used to rectify over-complete action instances in pseudo-labels, aligning them with corrected instances. Adjacent missing instances are compensated through MIC. The High-Quality Pseudo-Label Mining Loss (HPM), combined with the associated instance set, adaptively mines high-quality pseudo-labels, assigning greater optimization weights to these labels.

\subsection{Noisy Label Generator}
We first train the WTAL model based on video-level labels to obtain instance-level pseudo-labels. To refine the boundary of inaccurate instance of pseudo-labels, we propose CALA to enhance localization through context augmentation. The detail is shown as follows.
\subsubsection{WTAL Base Model}\label{WTAL Base Model}

Following the previous work \cite{pardo2021refineloc, he2022asm}, for each video $V_i$, we use the features $F \in R^{2048}$ extracted by the I3D network \cite{Carreira_Zisserman_2017}, where the I3D network is pre-trained on the Kinetics400 dataset. Using video-level labels, we train a WTAL via which we can obtain a noisy pseudo-label for each video sample. In this paper, the WTAL we use is ASM-Loc \cite{he2022asm} if not specified. Note that our method is decoupled from the WTAL model and can be easily cooperated to other weakly-supervised models. In the experiment section, we conduct an analysis of the generalizability for WTAL models.\par

\subsubsection{Context-Aware Label Augmentation}
\label{Context-aware based Label Augmentation}
Due to the sparsity of video-level label information, there exists non-negligible noises in the obtained pseudo-labels.


We propose the Context-Aware Label Augmentation (CALA) algorithm to refine instance pseudo-labels and reduce noise by considering neighboring instances. The CALA process is outlined in Algorithm \ref{algorithm-Context-Aware Label Augmentation}.

\begin{algorithm}[t]
\label{algorithm1}
    \caption{Context-Aware Label Augmentation (CALA)}
    \textbf{Input}: Weakly-supervised output $\mathcal{P}_{w}$,
    NMS threshold $\rho$, Confidence threshold $\psi$, IoU threshold $\eta_0$, Weight Aggregation Function $\mathcal{F}_{fuse}^{aug}(\cdot$, $\cdot)$
    \begin{algorithmic}[1] 
        \STATE $\mathcal{A}^{aug}, \mathcal{W}^{aug} = \phi$
        \STATE $\mathcal{P}_{filt}^{aug}=\{p_{t}|p_t \in \mathcal{P}_{w} \; \& \; {p_t}(\theta_{p_t}) > \psi\}$
        \STATE $index_{keep}^{aug}$=NMS$(\mathcal{P}_{filt}^{aug},\rho)$ 
        \STATE $\mathcal{P}_{keep}^{aug}=\mathcal{P}_{filt}^{aug}[index_{keep}^{aug},:]$
        \FOR{every $P_{a} \in \mathcal{P}_{keep}^{aug}$}
            \STATE  $\mathcal{P}_{a, ass},\mathcal{W}_{a,ass} = \phi$ 
            \FOR{every $P_w \in \mathcal{P}_{w}$} 
                \IF{$P_{a}(c_{a})=P_w(c_w)$ and IoU($P_{a},P_{w}$)$>\eta_0$} 
                 \STATE \# every associated action instance of $P_{a}$
                 \STATE Calculate adaptive weights $w_{a,w}$
                 \STATE $\mathcal{P}_{a, ass}=\mathcal{P}_{a, ass} \cup P_w$, $\mathcal{W}_{a,ass}=\mathcal{W}_{a,ass} \cup w_{a,w}$
                \ENDIF
            \ENDFOR
        \IF{$\mathcal{P}_{a, ass} \text{is not} \; \phi$}
          \STATE $P_{a}^{aug},w_{a}^{aug}=\mathcal{F}_{fuse}^{aug}(\mathcal{P}_{a, ass},\mathcal{W}_{a,ass}$)
       \ENDIF
        \STATE $\mathcal{A}^{aug}=\mathcal{A}^{aug} \cup P_{a}^{aug}$, $\mathcal{W}^{aug}=\mathcal{W}^{aug} \cup w_{a}^{aug}$
        \ENDFOR
        \RETURN Augmented Pseudo Ground-Truth $\mathcal{A}^{aug}$ and Weight $\mathcal{W}^{aug}$.
    \end{algorithmic}
    \label{algorithm-Context-Aware Label Augmentation}
\end{algorithm}

First, initialize the enhanced pseudo-labels and their corresponding weights. By setting a confidence threshold and deploying NMS processing, a set of high-confidence and low-overlapping action instances $P_{keep}^{aug}$ is selected. The context enhancement process is executed for each action instance $P_a$ in $\mathcal{P}_{keep}^{aug}$, including determining the associated action instance set $\mathcal{P}_{a,ass}$ and calculating the adaptive weights. The constraint for the associated action instance is: a) sharing the same category; b) having an IoU greater than a given threshold $\eta_0$. The formula for calculating the adaptive weight of $P_a$'s associate instance $P_r$ is as follows:

\begin{equation}
    \label{adaptive weight1}
w_r = e^{\sqrt{\text{IoU}(P_r, \mathcal{P}_{keep}^{aug}) \cdot \text{min}(\text{max}(P_r(\theta_r),0),1)}},
\end{equation}

where $P_r(\theta_r)$ is the confidence score of $P_r \in \mathcal{P}_{a,ass}$. If $\mathcal{P}_{a,ass}$ is not empty, then the augmented $P_a^{aug}$ of $P_a$ is obtained via weighted aggregation of associated action instances set $\mathcal{P}_{a,ass}$, which is computed as follows:

\begin{equation}
    \begin{aligned}
         P_a^{aug}(\zeta_a) & = \mathcal{F}_{fuse,stage}^{aug}(P_a, \mathcal{P}_{a,ass}) \\
         & = \frac{\sum_{i=1}^n w_i \cdot P_i(\zeta_i), P_i \in \mathcal{P}_{a,ass}}{\sum_{i=1}^n w_i},\\
    \end{aligned}
\end{equation}

\begin{equation}
    \text{where} \; \zeta_i=\left\{
\begin{array}{rcl}
s_i       &      & stage = start\\
e_i     &      & stage = end,\\
\end{array} \right.
\end{equation}

weight aggregation function $\mathcal{F}_{fuse}^{aug}(\cdot$, $\cdot)$ consists of two stage: start timestamp aggregation $\mathcal{F}_{fuse, start}^{aug}(\cdot$, $\cdot)$ and end timestamp aggregation $\mathcal{F}_{fuse, end}^{aug}(\cdot$, $\cdot)$. 
The Weight of $P_a^{aug}$ is computed as
$
    w_a^{aug} = \sqrt[n]{\prod_{i=1}^n w_i}.
$
Then add the $P_a^{aug}$ to the augmented pseudo-label set $\mathcal{A}^{aug}$ and repeat this process till all action instance is scanned.

\subsection{Online Noise Correction Framework}
The online noise correction framework aims to progressively correct three types of noise: missing action instances, many-to-one and incomplete instance problem. Specifically, we introduce a teacher-student-based online noise correction framework. 
As illustrated in Figure \ref{pipeline}, our online noise correction framework consists of several components: feature extraction encoder, data augmentation module, student model, teacher model, AIC, MIC, online pseudo-label set $\mathcal{A}^*$ and HPM.

\subsubsection{Teacher Model Pre-training}
The online noise correction process begins with the pre-training of the teacher model, involving using the enhanced pseudo-labels generated by CALA for full supervised training. After completing pre-training, the teacher model plays a crucial role in the online generation of stable compensation information during the noise correction stage, aiding the processes of AIC and MIC. The online pseudo-label $\mathcal{A}^*$ and weight $\mathcal{W}^*$ is first initialized as $\mathcal{A}^{aug}$ and $\mathcal{W}^{aug}$.
\subsubsection{Ambiguous Instances Correction}
\label{AIC text}
Ambiguous Instances Correction (AIC) aims to leverage the pseudo-label information derived from teacher predictions to accurately position instance boundaries through adjacent context aggregation, thus enabling the online update of the pseudo-label set. For low-quality samples, the output from the teacher model is further corrected through the AIC module, effectively suppressing noise in the pseudo-labels. Compared with traditional semi-supervised learning, we possess not only the output of the teacher model but also maintain an online, real-time corrected pseudo-label set. We refer to \textbf{Supplementary} for the workflow of AIC. \par

The input to the AIC consists of teacher predictions $\mathcal{P}_w$ and the online pseudo-label set $\mathcal{A}^*$. For each instance $A_r^*$ in the $\mathcal{A}^*$, we calculate its IoU with every action instance $P_t \in \mathcal{P}_w$. If the IoU is greater than the threshold $\eta_1$, and the confidence score of the action instance exceeds the threshold $\psi$, we include it in the corresponding associated set $\mathcal{A}_{r,ass}$ and compute its adaptive weight. The formula for calculating the adaptive weight $w_t$ of $A_r^*$'s associated instance $P_t$ is the same as the one proposed in the CALA algorithm, as described above in equation (\ref{adaptive weight1}). By combining $\mathcal{A}_{r,ass}$ and $A_r^*$, we perform weighted aggregation to obtain the corrected action instance $A_{r}^{cor}$ of $A_r^*$ as follows:
\begin{equation}
\begin{aligned}
     A_r^{cor}(\zeta_r) 
     & = \mathcal{F}_{fuse, stage}^{cor}(A_r^*, \mathcal{A}_{r,ass})
     = \alpha \cdot w_r^* \cdot A_r^*(\zeta_r) \\
     & + (1-\alpha) \cdot  \frac{\sum_{i=1}^n w_i \cdot A_i(\zeta_i), A_i \in \mathcal{A}_{r,ass}}{\sum_{i=1}^n w_i},
\end{aligned}
\end{equation}
\begin{equation}
    \text{where} \; \zeta_i=\left\{
\begin{array}{rcl}
s_i       &      & stage = start\\
e_i     &      & stage = end,\\
\end{array} \right.
\end{equation}
where $\alpha$ is hyper-parameter, $w_{r}^*$ is the weight of instance $A_r^*$'s weight in $\mathcal{W}^*$, $n$ is the size of $A_{r,ass}$.The corrected weight of $A_r^*$ is computed as:
\begin{equation}\label{AIC weight}
    w_r^{cor} = \sqrt[n+\beta]{(\prod_{i=1}^n w_i) \cdot (w_r^*)^\beta}, \; \;  A_i \in \mathcal{A}_{r,ass},
\end{equation}
where $\beta$ is hyper-parameter. Then add the corrected instance $A_r^{aug}$ of $A_r^*$ to the corrected pseudo-label set $\mathcal{A}^{cor}$, and add $ w_r^{cor}$ to corrected weight set $\mathcal{W}^{cor}$. \par
When all instance in $\mathcal{A}^*$ is corrected, the online set $\mathcal{A}^*$ is updated as $\mathcal{A}^{cor}$. The online weight $\mathcal{W}^*$ is updated as $\mathcal{W}^{cor}$. Through this updating, the $\mathcal{A}^*$ and $\mathcal{W}^*$ is corrected.

\subsubsection{Missing Instance Compensation}
\label{MIC text}
MIC aims to further update the pseudo-label set online, leveraging the potential missing information in neighboring pseudo-label instances, which are generated based on teacher predictions. Although AIC effectively utilizes contextual information within the pseudo-labels, it only focuses on correcting existing action instances in noisy pseudo-labels and does not address the task of compensating missing action instances. We refer to \textbf{Supplementary} for the main workflow of the MIC. \par
MIC takes teacher predictions $\mathcal{P}_t$ and the online pseudo-label set $\mathcal{A}^*$ as inputs. First, MIC filters out action instances with low confidence from $\mathcal{P}_t$, saving as $\mathcal{P}_{filt}^{com}$. Then it applies NMS to select low IoU instance on $\mathcal{P}_{filt}^{com}$ to obtain $\mathcal{P}_{keep}^{com}$. Next, for each instance $P_n$ in $\mathcal{P}_{keep}^{com}$, MIC computes the IoU with every action instance in $\mathcal{A}^*$. If the maximum IoU is less than $\eta_2$, $P_n$ is included in the compensated pseudo-label set $\mathcal{A}^{com}$. The adaptive weight $w_{n}^{com}$ of $A_n \in \mathcal{A}^{com}$ is obtained through:
\begin{equation}
\label{MIC weight}
    w_{n}^{com} = e^{\text{min}(\text{max}(A_n(\theta_n),0),1)},
\end{equation}
where $A_n(\theta_n)$ is the confidence score of $A_n$. $w_{n}^{com}$ is added into compensated weight $\mathcal{W}^{com}$.\par
Through the MIC module, missing action instances are filled. Similar before, we update $A^*$ as $\mathcal{A}^{com}$, $\mathcal{W}^*$ as $\mathcal{W}^{com}$ again to achieve online label compensation.

\subsubsection{High-Quality Pseudo-Label Mining}
\label{High-Quality text}

To fully utilize the supervision information from accurate pseudo-labels, we introduce adaptive weights for each action instance by assigning higher optimization weights to high-quality instances and vice versa. 
Specially, we propose the High-Quality Pseudo-Label Mining (HPM) Loss, which is formulated as:
\begin{equation}
\begin{aligned}
    L 
    & = \frac{\lambda}{N_{pos}} \sum_{l, t}\mathds{1}_{\{c_t^l>0\}} \cdot w_t \cdot (\sigma_{IoU} L_{cls} + L_{reg}) \\
    & + \frac{1}{N_{reg}} \sum_{l,t} \mathds{1}_{\{c_t^l=0\}}\cdot L_{cls},
\end{aligned}
\end{equation}
where $\lambda$ represents the weight of positive sample loss, $N_{pos}$ and $N_{neg}$ denote the number of positive and negative samples, and $l$ stands for the number of classification heads and Trident-head heads. For each classification head and Trident-head, we compute foreground classification loss $L_{cls}$, regression loss $L_{reg}$, and background classification loss sequentially. In this paper, $L_{cls}$ is focal loss and $L_{reg}$ is IoU loss. It's worth noting that using adaptive weights $w_t$ for each foreground action instance gives more importance to high-quality instances during the optimization process.

\begin{table*}[htb]
    \centering
    \resizebox{\textwidth}{!}{
\begin{tabular}{c|c|ccccccc|ccc}
\hline
& & \multicolumn{7}{c|}{\textbf{mAP@IoU(\%)}} & \multicolumn{3}{c}{\textbf{AVG mAP(\%)}}                                                                             \\ \cline{3-12} 
\multirow{-2}{*}{\textbf{Sup.}} & \multirow{-2}{*}{\textbf{Method}}  & 0.1   & 0.2   & 0.3   & 0.4   & 0.5   & 0.6   & 0.7   & 0.1:0.5   & 0.3:0.7   & 0.1:0.7                               \\ \hline
& SSN \cite{zhao2017temporal}& 60.3  & 56.2  & 50.6  & 40.8  & 29.1  & - & - & 47.4  & - & -                                     \\
\multirow{-2}{*}{\textbf{Fully}}& TAL-Net\cite{chao2018rethinking}  &  59.8  & 57.1  & 53.2  & 48.5  & 42.8  & 33.8  & 20.8  & 52.3  & 39.8  & 45.1                                  \\ \hline
& SF-Net  \cite{ma2020sf}   &  68.3  & 62.3  & 52.8  & 42.2  & 30.5  & 20.6  & 12.0  & 51.2  & 31.6  & 41.2                                  \\
& DCM  \cite{ju2021divide} &  70.2  & 63.5  & 55.6  & 44.7  & 32.3  & 22.0  & 12.3  & 53.3  & 33.4  & 42.9                                  \\
& BackTAL\cite{yang2021background}  & - & - & 54.4  & 45.5  & 36.3  & 26.2  & 14.8  & - & 35.4  & -                                     \\
& PCL \cite{li2023prototype}& 74.6  & 70.2  & 63.3  & 55.9  & 44.4  & - & - & 61.7  & - & -                                     \\ \cline{2-12} 
& \cellcolor[HTML]{D7D7D7}LACP \cite{lee2021learning}& \cellcolor[HTML]{D7D7D7} \cellcolor[HTML]{D7D7D7}75.7  & \cellcolor[HTML]{D7D7D7}71.4  & \cellcolor[HTML]{D7D7D7}64.6  & \cellcolor[HTML]{D7D7D7}56.5  & \cellcolor[HTML]{D7D7D7}45.3  & \cellcolor[HTML]{D7D7D7}34.5  & \cellcolor[HTML]{D7D7D7}21.8  & \cellcolor[HTML]{D7D7D7}62.7  & \cellcolor[HTML]{D7D7D7}44.5  & \cellcolor[HTML]{D7D7D7}52.8          \\
& \cellcolor[HTML]{D7D7D7}\textbf{LACP+NoCo}  & \cellcolor[HTML]{D7D7D7}78.8  & \cellcolor[HTML]{D7D7D7}74.6  & \cellcolor[HTML]{D7D7D7}68.9  & \cellcolor[HTML]{D7D7D7}60.1  & \cellcolor[HTML]{D7D7D7}49.6  & \cellcolor[HTML]{D7D7D7}37.2  & \cellcolor[HTML]{D7D7D7}24.2  & \cellcolor[HTML]{D7D7D7}66.4  & \cellcolor[HTML]{D7D7D7}48.0  & \cellcolor[HTML]{D7D7D7}56.2          \\
& \cellcolor[HTML]{D7D7D7}HR-Pro \cite{zhang2024hr}  & \cellcolor[HTML]{D7D7D7} \cellcolor[HTML]{D7D7D7}85.6  & \cellcolor[HTML]{D7D7D7}81.6  & \cellcolor[HTML]{D7D7D7}74.3  & \cellcolor[HTML]{D7D7D7}64.3  & \cellcolor[HTML]{D7D7D7}52.2  & \cellcolor[HTML]{D7D7D7}39.8  & \cellcolor[HTML]{D7D7D7}24.8  & \cellcolor[HTML]{D7D7D7}71.6  & \cellcolor[HTML]{D7D7D7}51.1  & \cellcolor[HTML]{D7D7D7}60.4          \\
\multirow{-8}{*}{\textbf{Point-level}} & \cellcolor[HTML]{D7D7D7}\textbf{HR-Pro+NoCo}   & \cellcolor[HTML]{D7D7D7}\textbf{86.0}  & \cellcolor[HTML]{D7D7D7}\textbf{82.1}  & \cellcolor[HTML]{D7D7D7}\textbf{75.6}  & \cellcolor[HTML]{D7D7D7}\textbf{65.7}  & \cellcolor[HTML]{D7D7D7}\textbf{53.6}  & \cellcolor[HTML]{D7D7D7}\textbf{40.6}  & \cellcolor[HTML]{D7D7D7}\textbf{25.3}  & \cellcolor[HTML]{D7D7D7}\textbf{72.6}  & \cellcolor[HTML]{D7D7D7}\textbf{52.2}  & \cellcolor[HTML]{D7D7D7}\textbf{61.3}          \\ \hline
& CoLA \cite{zhang2021cola} &  66.2  & 59.5  & 51.5  & 41.9  & 32.2  & 22.0  & 13.1  & 50.3  & 32.1  & 40.9                                  \\
& DELU \cite{chen2022dual}   &  71.5  & 66.2  & 56.5  & 47.7  & 40.5  & 27.2  & 15.3  & 56.5  & 37.4  & 46.4                                  \\
& Zhou et al. \cite{zhou2023improving} & 74.0  & 69.4  & 60.7  & 51.8  & 42.7  & 26.2  & 13.1  & 59.7  & 38.9  & 48.3                                  \\
& Ju et al. \cite{ju2023distilling}&  73.5  & 68.8  & 61.5  & 53.8  & 42.0  & 29.4  & 16.8  & 59.9  & 40.7  & 49.4                                  \\
& PivoTAL \cite{rizve2023pivotal} &  74.1  & 69.6  & 61.7  & 52.1  & 42.8  & 30.6  & 16.7  & 60.1  & 40.8  & 49.7                                  \\
&  CASE   \cite{liu2023revisiting}   &  72.3  & 67.1  & 59.2  & 49.4  & 37.7  & 24.2  & 13.7  & 57.1  & 36.8  & 46.2                                  \\
& ISSF \cite{yun2024weakly}&  72.4  & 66.9  & 58.4  & 49.7  & 41.8  & 25.5  & 12.8  & 57.8  & 37.6  & 46.8                                  \\ \cline{2-12} 
& \cellcolor[HTML]{D7D7D7}{ {CO$_2$-Net \cite{hong2021cross}}} &  \cellcolor[HTML]{D7D7D7}{70.1} & \cellcolor[HTML]{D7D7D7}{63.6} & \cellcolor[HTML]{D7D7D7}{54.5} & \cellcolor[HTML]{D7D7D7}{45.7} & \cellcolor[HTML]{D7D7D7}{38.3} & \cellcolor[HTML]{D7D7D7}{26.4} & \cellcolor[HTML]{D7D7D7}{13.4} & \cellcolor[HTML]{D7D7D7}{54.4} & \cellcolor[HTML]{D7D7D7}{35.7} & \cellcolor[HTML]{D7D7D7}{44.6} \\
& \cellcolor[HTML]{D7D7D7}{\textbf{CO$_2$-Net+NoCo}}    & \cellcolor[HTML]{D7D7D7}{74.6} & \cellcolor[HTML]{D7D7D7}{70.3} & \cellcolor[HTML]{D7D7D7}{63.0} & \cellcolor[HTML]{D7D7D7}{53.9} & \cellcolor[HTML]{D7D7D7}{44.0} & \cellcolor[HTML]{D7D7D7}{31.7} & \cellcolor[HTML]{D7D7D7}{17.4} & \cellcolor[HTML]{D7D7D7}{61.2} & \cellcolor[HTML]{D7D7D7}{42.0} & \cellcolor[HTML]{D7D7D7}{50.7} \\
& \cellcolor[HTML]{D7D7D7}{ {ASM-Loc \cite{he2022asm}}}    & \cellcolor[HTML]{D7D7D7}{71.2} & \cellcolor[HTML]{D7D7D7}{65.5} & \cellcolor[HTML]{D7D7D7}{57.1} & \cellcolor[HTML]{D7D7D7}{46.8} & \cellcolor[HTML]{D7D7D7}{36.6} & \cellcolor[HTML]{D7D7D7}{25.2} & \cellcolor[HTML]{D7D7D7}{13.4} & \cellcolor[HTML]{D7D7D7}{55.4} & \cellcolor[HTML]{D7D7D7}{35.8} & \cellcolor[HTML]{D7D7D7}{45.1} \\
\multirow{-11}{*}{\textbf{Video-level}} & \cellcolor[HTML]{D7D7D7}{\textbf{ASM-Loc+NoCo}}    & \cellcolor[HTML]{D7D7D7}{\textbf{75.2}} & \cellcolor[HTML]{D7D7D7}{\textbf{70.7}} & \cellcolor[HTML]{D7D7D7}{\textbf{63.0}} & \cellcolor[HTML]{D7D7D7}{\textbf{54.1}} & \cellcolor[HTML]{D7D7D7}{\textbf{44.0}} & \cellcolor[HTML]{D7D7D7}{\textbf{31.7}} & \cellcolor[HTML]{D7D7D7}{\textbf{17.7}} & \cellcolor[HTML]{D7D7D7}{\textbf{61.4}} & \cellcolor[HTML]{D7D7D7}{\textbf{42.1}} & \cellcolor[HTML]{D7D7D7}{\textbf{50.9}} \\ \hline
\end{tabular}
    }
        \caption{Comparisons with state-of-the-art methods. AVG is the average mAP under the IoU thresholds 0.1:0.7:0.1 for THUMOS14 dataset. ’-’ means that the corresponding results are nor reported in the original papers.}
    \label{new-table1}
    \end{table*}

\begin{table}[htb]
\centering
\renewcommand\arraystretch{1.2}
\resizebox{\linewidth}{!}{
\begin{tabular}{c|c|cccc}

\hline
& &  \multicolumn{4}{c}{\textbf{mAP@IoU(\%)}}                                                                                 \\ \cline{3-6} 
\multirow{-2}{*}{\textbf{Sup.}} & \multirow{-2}{*}{\textbf{Method}}    & 0.5  & 0.75 & 0.95& AVG                          \\ \hline
\multirow{-1}{*}{\textbf{Fully}}   & SSN \cite{zhao2017temporal} & 41.3 & 27.0 & 6.1 & 26.6                         \\ \hline
& SF-Net  \cite{ma2020sf}& 37.8 & -& -   & 22.8                         \\ \cline{2-6} 
& \cellcolor[HTML]{D7D7D7}BackTAL \cite{yang2021background}&  \cellcolor[HTML]{D7D7D7}41.5 & \cellcolor[HTML]{D7D7D7}27.3 & \cellcolor[HTML]{D7D7D7}4.7 & \cellcolor[HTML]{D7D7D7}27.0 \\
& \cellcolor[HTML]{D7D7D7}\textbf{BackTAL+NoCo}   & \cellcolor[HTML]{D7D7D7}49.8 & \cellcolor[HTML]{D7D7D7}31.6 & \cellcolor[HTML]{D7D7D7}6.1 & \cellcolor[HTML]{D7D7D7}30.8 \\
& \cellcolor[HTML]{D7D7D7}LACP \cite{lee2021learning}  &  \cellcolor[HTML]{D7D7D7}44.0 & \cellcolor[HTML]{D7D7D7}26.0 & \cellcolor[HTML]{D7D7D7}5.9 & \cellcolor[HTML]{D7D7D7}26.8 \\
\multirow{-5}{*}{\textbf{Point-level}}  & \cellcolor[HTML]{D7D7D7}\textbf{LACP+NoCo}   &  \cellcolor[HTML]{D7D7D7}\textbf{50.2} & \cellcolor[HTML]{D7D7D7}\textbf{30.1} & \cellcolor[HTML]{D7D7D7}\textbf{6.8} & \cellcolor[HTML]{D7D7D7}\textbf{30.7} \\ \hline
&CoLA \cite{zhang2021cola}  &  42.7 & 25.7 & 5.8 & 26.1                         \\
& DELU \cite{chen2022dual}   &  44.2 & 26.7 & 5.4 & 26.9                         \\
& Ren et al. \cite{ren2023proposal}  &  44.2 & 26.1 & 5.3 & 25.5                         \\
& Ju et al. \cite{ju2023distilling}  &  48.3 & 29.3 & 6.1 & 29.6                         \\
& CASE  \cite{liu2023revisiting}  &  43.8 & 27.2 & 6.7 & 27.9                         \\
& \cellcolor[HTML]{D7D7D7}{CO$_2$-Net \cite{hong2021cross}}   & \cellcolor[HTML]{D7D7D7} \cellcolor[HTML]{D7D7D7}43.3 & \cellcolor[HTML]{D7D7D7}26.3 & \cellcolor[HTML]{D7D7D7}5.2 & \cellcolor[HTML]{D7D7D7}26.4 \\ 
& \cellcolor[HTML]{D7D7D7}\textbf{CO$_2$-Net+NoCo}   & \cellcolor[HTML]{D7D7D7}48.9 & \cellcolor[HTML]{D7D7D7}30.0 & \cellcolor[HTML]{D7D7D7}6.6 & \cellcolor[HTML]{D7D7D7}30.1 \\ 
& \cellcolor[HTML]{D7D7D7}{ASM-Loc \cite{he2022asm}} & \cellcolor[HTML]{D7D7D7} \cellcolor[HTML]{D7D7D7}43.4 & \cellcolor[HTML]{D7D7D7}26.6 & \cellcolor[HTML]{D7D7D7}5.3 & \cellcolor[HTML]{D7D7D7}26.5 \\ 
\multirow{-9}{*}{\textbf{Video-level}} & \cellcolor[HTML]{D7D7D7}\textbf{ASM-Loc+NoCo}  & \cellcolor[HTML]{D7D7D7}\textbf{49.1} & \cellcolor[HTML]{D7D7D7}\textbf{30.5} & \cellcolor[HTML]{D7D7D7}\textbf{6.7} & \cellcolor[HTML]{D7D7D7}\textbf{30.7} \\ \hline
\end{tabular}
}
\caption{Detection results on ActivityNet1.2}
\label{new-table2}
\end{table}

\begin{table}[h]
\renewcommand\arraystretch{1.2}
\resizebox{\linewidth}{!}{
\begin{tabular}{c|c|c|c|c|c|c}
\hline
\multirow{1}{*}{\#}& \multirow{1}{*}{\textbf{CALA}}& \multirow{1}{*}{\textbf{T-S}}& \multirow{1}{*}{\textbf{AIC}}& \multirow{1}{*}{\textbf{MIC}}& \multirow{1}{*}{\textbf{HPM}}& \multicolumn{1}{c}{\textbf{AVG mAP(\%)}}                                                 \\ \hline
a&&&&&&  35.8 \\ \hline
b& $\checkmark$&&&&&  37.1 \textcolor{red}{(+1.3)} \\ \hline
c& $\checkmark$& $\checkmark$&&&& 37.8 \textcolor{red}{(+2.0)} \\ \hline
d& $\checkmark$& $\checkmark$&$\checkmark$&&&  40.0 \textcolor{red}{(+4.2)} \\ \hline
e& $\checkmark$& $\checkmark$&&$\checkmark$&&  39.6 \textcolor{red}{(+3.8)} \\ \hline
f& $\checkmark$& $\checkmark$& $\checkmark$& $\checkmark$&&  41.9 \textcolor{red}{(+6.1)} \\ \hline
g& $\checkmark$& $\checkmark$& $\checkmark$& $\checkmark$& $\checkmark$&  42.1 \textcolor{red}{(+6.3)} \\ \hline
\end{tabular}
}
\caption{Ablation study of core components,'T-S' denotes the Teacher-Student module.}
\label{Ablation-table}
\end{table}

\begin{table}[t]
\centering
\renewcommand\arraystretch{1.2}
\resizebox{\linewidth}{!}{
\begin{tabular}{c|ccccc|c}
\hline
\multirow{1}{*}{\textbf{WTAL}}& \multirow{1}{*}{\textbf{CALA}}& \multirow{1}{*}{\textbf{T-S}}& \multirow{1}{*}{\textbf{MIC}}& \multirow{1}{*}{\textbf{AIC}}& \multirow{1}{*}{\textbf{HPM}}& \multirow{1}{*}{\textbf{AVG mAP(\%)}}                      \\ \hline
\multirow{6}{*}{UM}&&&&&&  30.3          \\
& $\checkmark$&&&&&  31.6 \textcolor{red}{(+1.3)}          \\
& $\checkmark$& $\checkmark$&&&&  32.3    \textcolor{red}{(+2.0)}    \\
& $\checkmark$& $\checkmark$& $\checkmark$&&&  33.9  \textcolor{red}{(+3.6)}         \\
& $\checkmark$& $\checkmark$& $\checkmark$& $\checkmark$&&  36.1   \textcolor{red}{(+5.8)}       \\
& $\checkmark$& $\checkmark$& $\checkmark$& $\checkmark$& $\checkmark$& \textbf{36.4 \textcolor{red}{(+6.1)}}  \\ \hline
\multirow{6}{*}{CO$_2$-Net} &             &             &             &&&  36.0            \\
& $\checkmark$&             &             &             &             & 37.0    \textcolor{red}{(+1.0)}       \\
& $\checkmark$& $\checkmark$&             &             &             & 37.7    \textcolor{red}{(+1.7)}       \\
& $\checkmark$& $\checkmark$& $\checkmark$&             &             & 40.2    \textcolor{red}{(+4.2)}      \\
& $\checkmark$& $\checkmark$& $\checkmark$& $\checkmark$&             & 41.8    \textcolor{red}{(+5.8)}      \\
& $\checkmark$& $\checkmark$& $\checkmark$& $\checkmark$& $\checkmark$& \textbf{42.0 \textcolor{red}{(+6.0)}}   \\ \hline
\end{tabular}
}
\caption{Performances under two typical WTAL algorithms on THUMOS14 dataset.}
\label{generalized WS}
\end{table}

\section{Experiment}
\subsection{Settings and Datasets}
\textbf{Datasets.} We evaluate our method on THUMOS14 and ActivityNet v1.2, following \cite{yang2021multi}. 
\textbf{THUMOS14} contains 200 validation and 213 test samples. The validation set is used for training, and the test set for evaluation.  
\textbf{ActivityNet v1.2} includes 4,819 training, 2,383 validation, and 2,489 test videos across 100 action classes, averaging 1.5 instances per video. We use the training set for training and the validation set for evaluation.

\noindent\textbf{Implementation.}We use the I3D model for feature extraction and ASM-Loc \cite{he2022asm} as the default noise label generator unless stated otherwise. Additional dataset-specific details are in the supplementary material. We follow standard protocols with IoU thresholds to measure localization accuracy and report mIoU for pseudo-label quality across foreground and background categories.

\subsection{Comparison with State-Of-Art Methods}
We compare NoCo with the existing methods on ActivityNet v1.2 and THUMOS14.The result is shown in Table \ref{new-table1}.\par

\noindent\textbf{Results on THUMOS14.}
NoCo outperforms all WTAL baseline methods. Its average mAP is 6.1\% higher than CO2-Net and 5.8\% higher than ASM-Loc . Additionally, NoCo surpasses the latest methods CASE , AHIM , DDG-Net , and the current leading method PivoTAL, pushing the average mAP to 50.9\%, setting a new state-of-the-art performance. Even when compared with FTAL methods, NoCo outperforms TAL-Net at all IoUs. Furthermore, we also applied NoCo to point-supervised temporal action localization methods LACP and HR-Pro, achieving 3.4\% and 0.9\% average mAP improvements over the baselines, respectively, resulting in current state-of-the-art performance.

\noindent\textbf{Results on ActivityNet v1.2.}
Our method outperforms existing methods across all metrics. Compared to the baseline ASM-Loc, our method improves mAP by 4.2\%, and compared to the baseline CO2-Net, it improves by 3.7\%. NoCo also surpasses Ju et al.\cite{ju2023distilling}, achieving a new state-of-the-art localization accuracy of 30.7\%. Our method even outperforms some previous fully supervised methods, such as SSN, demonstrating the feasibility of noise correction modeling. Additionally, we applied NoCo to point-supervised temporal action localization methods BackTAL and LACP, achieving average mAP improvements of 3.8\% and 3.9\%, respectively, compared to the baseline.\par


The consistent results on ActivityNet v1.2 and THUMOS14 demonstrate the effectiveness of our NoCo.\par

\subsection{Ablation Studies}

We conducted a series of ablation experiments to evaluate the contributions of each component in our method. The results, presented in Table \ref{Ablation-table}, are labeled from \textit{a} to \textit{g} for clarity.

\noindent\textbf{Baseline.}  
\textit{Experiment a} combines the weakly supervised model ASM-Loc with the fully supervised TriDet model for pseudo-label retraining on the THUMOS14 dataset, without any correction modules. Pseudo-labels were generated by filtering action instances with confidence scores above 0.5. However, this threshold-based method introduced substantial noise, resulting in minimal performance improvement.

\begin{figure*}[t]
	\centering
	\includegraphics[width=1.0\linewidth,height=3.5cm]{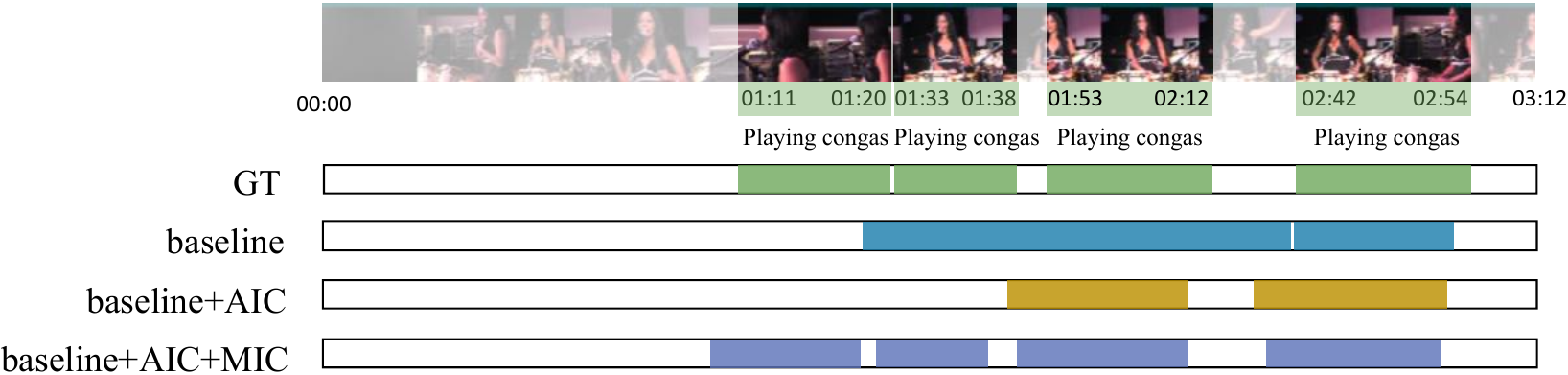}
	\caption{Visualization of ground-truth and predictions.}
	\label{fig:qua}
\end{figure*}

\noindent\textbf{Influence of CALA.}  
\textit{Experiment b} applies CALA, achieving a 1.3\% mAP increase, demonstrating its ability to enhance pseudo-labels with contextual information.

\noindent\textbf{Influence of Customized Teacher-Student Training.}  
\textit{Experiment c} introduces a customized teacher-student framework without pseudo-label correction modules. The teacher’s labels guide the student, improving performance by 0.7\% due to the student model’s robustness from prior teacher-student iterations.

\noindent\textbf{Influence of AIC.}  
\textit{Experiment d} adds AIC, boosting performance by 2.2\%. AIC corrects inaccurately localized pseudo-labels, refining the student model’s learning.

\noindent\textbf{Influence of MIC.}  
\textit{Experiment e} incorporates MIC, resulting in a 1.8\% improvement. MIC compensates for missing action instances in pseudo-labels, enhancing learning.

\noindent\textbf{Influence of Combined MIC and AIC.}  
\textit{Experiment f} applies both MIC and AIC, yielding a 4.1\% performance increase over \textit{experiment c}. AIC corrects overly complete instances, while MIC fills in missing actions, addressing the one-to-many issue.

\noindent\textbf{Influence of High-Quality Pseudo-Label Mining.}  
\textit{Experiment g} adds high-quality pseudo-label mining loss, providing a 0.2\% performance boost and reaching the highest mAP of 42.1\%.

\subsection{Generalizability Analysis of WTAL Models}

Our noise correction framework is decoupled from weakly supervised models, making it applicable to any existing temporal action localization method. To evaluate its generalizability, we apply our framework to the UM \cite{lee2021weakly} and CO$_2$-Net \cite{hong2021cross} models, with results shown in Table \ref{generalized WS}. 

For the UM model, the average mAP on THUMOS14 is 30.3\%. The CALA module improves this by 1.3\%, and the Teacher-Student module contributes an additional 2.0\%. The MIC, AIC, and HPM modules further increase the performance by 3.6\%, 5.8\%, and 6.1\%, respectively.

Similarly, for CO$_2$-Net, the mAP starts at 36.0\%. The sequential introduction of the CALA, Teacher-Student, MIC, AIC, and HPM modules results in improvements of 1.0\%, 1.7\%, 4.2\%, 5.8\%, and 6.0\%, respectively.

These results highlight the strong generalizability of our noise correction framework across various weakly-supervised models.

\subsection{Analysis of Inference Speed}


NoCo gains a speed advantage by running only the student model during inference. We compare the inference performance and speed in Table \ref{inference-table}. It can be observed that NoCo is not only faster than the baseline WTAL method, showing a threefold improvement, but also achieves more accurate localization results.


\begin{table}[t]
\centering
\renewcommand\arraystretch{1.2}
\resizebox{\linewidth}{!}{

\begin{tabular}{c|c|c|c}
\hline
                  & \textbf{mAP(0.3:0.7)↑} & \textbf{speed(video/s)↑} & \textbf{FLOPS↓} \\ \hline
\textbf{Baseline} & 35.8                   & 1.64                     & 124.79G         \\
\textbf{NoCo}     & 42.1                   & 6.67                     & 20.12G          \\ \hline
\end{tabular}

}
\caption{NoCo outperforms existing methods in terms of both localization average mAP and inference speed.}
\label{inference-table}
\end{table}

\subsection{Qualitative Results}
Figure \ref{fig:qua}  illustrates the  visual comparison between the base model and NoCo for the action \textit{Playing congas}. 
We observe that when introducing AIC, the action boundary is more precise compared to the baseline. Furthermore, with the addition of MIC, missing action instances are compensated for, and many-to-one problem is solved. These results provide evidence for the effectiveness of proposed AIC and MIC.

\section{Conslusion}

In this paper, we propose a progressive framework for correcting noisy pseudo-labels. We introduce a CALA module to obtain more accurate action boundaries. Subsequently, we establish  an online noise correction framework based on a teacher-student training strategy. This framework incorporates the AIC and MIC module to tackle the instances-missing and many-to-one problem. Furthermore, we introduce an HPM loss to discover high-quality pseudo-labels. Our approach achieves state-of-the-art performance on THUMOS14 and ActivityNet v1.2, with a threefold improvement in inference speed.

\newpage
\section{Acknowledgments}

This work was supported by the National Key R\&D Program of China (2022YFB4701400/4701402), SSTIC Grant (KJZD20230923115106012,KJZD20230923114916032,\\GJHZ20240218113604008), Beijing Key Lab of Networked Multimedia and National Natural Science Foundation of China under Grant 62202302.

\bigskip

\bibliography{aaai25}

\end{document}


\onecolumn
Sec.~\ref{section1} covers the details of the Ambiguity Instance Correction Module. Sec.~\ref{section2} delves into the details of the Missing Instance Compensation Module. Sec.~\ref{section3} provides an overview of experimental details and hyperparameter analysis. Sec.~\ref{section4} introduces the inference pipeline. Sec.~\ref{section5} presents additional visualization results. Sec.~\ref{section6} discusses Broader effects.

\section{Ambiguous Instance Correction.}
\label{section1}

In Algorithm~\ref{Ambiguous Instances Correction}, we provide a detailed explanation of how AIC (Ambiguous Instance Correction) utilizes compensatory information from the teacher model to correct online pseudo-labels. Firstly, AIC takes inputs such as teacher predictions $\mathcal{P}_w$ and confidence thresholds $\psi$, online pseudo-label set $\mathcal{A}^{ {* }}$, IoU threshold $\eta_1$, and a weighted aggregation function $\mathcal{F}_{\text {fuse,stage }}^{\text {cor }}(\cdot, \cdot)$.\\

The process begins by initializing the corrected pseudo-labels $\mathcal{A}^{c o r}$ and their weights $\mathcal{W}^{\text {cor }}$ as empty sets. Subsequently, for each action instance in the online pseudo-label set $\mathcal{A}^{{* }}$, AIC explores a set of highly IoU action instances $\mathcal{A}_{r, a s s}$. The constraint for inclusion in this set is that the IoU must exceed a certain threshold $\eta_1$, and the confidence level must be greater than a specified value $\psi$. If the associated instance set is not empty, AIC utilizes it to correct the action instance, resulting in a corrected action instance.  This correction process is repeated for all action instances, leading to the corrected pseudo-labels. The objective is achieved by replacing the original labels $\mathcal{A}^{ {* }}$ with the corrected ones $\mathcal{A}^{c o r}$, thereby achieving online correction of pseudo-labels.
\renewcommand{\thealgorithm}{2}
\begin{algorithm}[!h]
\label{algorithm1}
    \caption{Ambiguous Instances Correction (AIC)}
    \textbf{Input}: Teacher Prediction $\mathcal{P}_w$,  Teacher Prediction Confidence $\Theta^{t}$,t online pseudo-label $\mathcal{A}^{*}$, Confidence Threshold $\psi$, IoU Thresholds $\eta_1$, Weight Aggregation Function $\mathcal{F}_{\text {fuse,stage }}^{\text {cor }}(\cdot, \cdot)$
    \begin{algorithmic}[1] 
        \STATE $\mathcal{A}^{cor}, \mathcal{W}^{cor} = \phi$ ~~~~~~~~~~~~~~~~~~~~~~~~~~~~~~~~~~~~~~~~~~~~~~~~~~~~~~`//Initialize the correction instance set to be empty
        \FOR{every $P_r \in \mathcal{A}^{{*}}$}                 
            \STATE $P_{r,ass},W_{r,ass} = \phi$ ~~~~~~~~~~~~~~~~~~~~~~~~~~~~~~~~~~~~~~~~~~~~~~~~//Initialize the association instance set.
            \FOR{every $P_w \in \mathcal{P}_w$}                 
                \IF{$P_w(c_w) = P_r(c_r)$ and IoU($P_w, P_r$)$ > \eta_1$ and $\Theta^{t}_{P_w}>\psi$}       
                 \STATE Get adaptive weights $w_{r, t}$=h($P_w, P_r$)\  ~~~~~//Constraints include the same category, high confidence, and large IoU
                 \STATE $\mathcal{A}_{r,ass}=\mathcal{A}_{r,ass} \cup \{(P_w)\},\mathcal{W}_{r,ass}=\mathcal{W}_{r,ass} \cup \{(w_{r, t})\}$
                \ENDIF
            \ENDFOR
            \IF{$\mathcal{A}_{a, ass} \text{is not} \; \phi$}   
                \STATE $P_{r}^{cor}$,$w_{r}^{cor}$=$\mathcal{F}_{\text {fuse,stage }}^{\text {cor }}(\mathcal{A}_{r,ass}$,$\mathcal{W}_{r,ass})$    ~~~~~~~~~~~~~~~~//Compute the corrected instance.
            \ENDIF
            \STATE $\mathcal{A}^{cor}=\mathcal{A}^{cor} \cup \{P_{r}^{cor}\}$, $\mathcal{W}^{cor}=\mathcal{W}^{cor} \cup \{w_{r}^{cor}\}$   ~//Add to the corrected pseudo-label.
        \ENDFOR
        \RETURN Corrected Pseudo Ground-Truth $P_{cor}$ and Weight $W_{cor}$.
    \end{algorithmic}
    \label{Ambiguous Instances Correction}
\end{algorithm}

\section{Missing Instance Compensation.}
\label{section2}
In Algorithm ~\ref{Ambiguous Instances Correction1}, we provide a detailed explanation of how MIC (Missing Instance Compensation) utilizes compensatory information from the teacher model to compensate for missing action instances in online pseudo-labels. Firstly, MIC takes inputs such as teacher predictions $\mathcal{P}_t$ and confidence threshold $\psi$, online pseudo-label $\mathcal{A}^{ {* }}$, Non-Maximum Suppression (NMS) threshold $\rho$, confidence threshold $\psi$, and IoU threshold $\eta_2$.

The process begins by initializing compensation pseudo-labels $\mathcal{A}^{c o m}$ and their weights $\mathcal{W}^{c o r}$. Then, a confidence filter is applied to obtain a set of instances with high confidence $\mathcal{P}^{c o m}_{filt}$. Subsequently, NMS processing is performed to obtain a set of instances $P_{\text {keep }}^{\text {com }}$ with low overlap. For each instance in this set, the $\mathrm{IoU}_{\mathrm{max}}$ with all instances is calculated. If $\mathrm{IoU}_{\mathrm{max}}$ is below a certain threshold $\eta_2$, a quality assessment weight is determined based on $w_n^{\text {com }}=\exp \left(\min \left(\max \left(A_n\left(\theta_n\right), 0\right), 1\right)\right)$ and added to the corresponding instance and its weight. This process is repeated for each instance in $P_{\text {keep }}^{\text {com }}$, resulting in the compensation pseudo-labels $\mathcal{A}^{c o m}$. The objective is achieved by replacing the original labels $\mathcal{A}^{c o m}$ with the compensated ones $\mathcal{A}^{c o m}$, thus achieving online compensation of pseudo-labels.
\renewcommand{\thealgorithm}{3}
\begin{algorithm}[t]
\label{algorithm1}
    \caption{Missing Instance Compensation (MIC)}
    \textbf{Input}: Teacher Prediction $\mathcal{P}_{t}$, Teacher Prediction Confidence $\Theta^{t}$, Augmented Pseudo Ground-Truth Set $\mathcal{A}^{*}$, NMS threshold $\rho$, Confidence Threshold $\psi$, IoU Thresholds $\eta_2$
    \begin{algorithmic}[1] 
        \STATE $\mathcal{A}^{com}, \mathcal{W}^{com} = \mathcal{A}^{*}, \mathcal{W}^{*} $    ~~~~~~~~~~~~~~~~~~~~~~~~~~~~~~~~~~~~~~~~~~~~~~~~~~~~~~~~~~~~~~~~~~~~~~//Initialize to empty
        \STATE $\mathcal{P}_{filter}^{com}=\{P_{t}|P_t \in \mathcal{P}_{t} \; \& \; \Theta^{t}_{P_t} > \psi\}$      ~~~~~~~~~~~~~~~~~~~~~~~~~~~~~~~~~~~~~~~~~~~~~~~~~~~//Remove instances with low confidence.
        \STATE $index_{keep}^{MIC}$=NMS$(\mathcal{P}_{filter}^{com},\rho)$ ~~~~~~~~~~~~~~~~~~~~~~~~~~~~~~~~~~~~~~~~~~~~~~~~~~~~~~~~~~~~~~~//Exclude overlapping instances.
        \STATE $\mathcal{P}_{keep}=\mathcal{P}_{filter}^{com}[index_{keep}^{MIC},:]$
        \FOR{every $P_n \in \mathcal{P}_{keep}^{com}$}
            \STATE $\mathcal{I}_{IoU} = \phi$
            \FOR{every $P_t \in \mathcal{P}^{*}$}
                \STATE $\mathcal{I}_{IoU}=\mathcal{I}_{IoU} \cup \{IoU(p_n,p_t)\}$
            \ENDFOR
            \STATE Compute maximum of IoU as $\text{IoU}_{max}$ 
            \IF{$\text{IoU}_{max} < \eta_2$}    
            \STATE Get adaptive weight $w_{n}^{com}$ \
            \STATE $\mathcal{A}^{com} = \mathcal{A}^{com} \cup \{P_n\},\mathcal{W}^{com} = \mathcal{W}^{com} \cup \{w_{n}^{com}\}$  ~~~~~~~~~~~~~~~~~~~~~//Add to the compensation pseudo-label.
            \ENDIF
        \ENDFOR
        \RETURN Compensated Pseudo Ground-Truth $\mathcal{A}_{com}$ and Weight $\mathcal{W}_{com}$.
    \end{algorithmic}
    \label{Ambiguous Instances Correction1}
\end{algorithm}
\section{Experiment Details.}
\label{section3}
During experiments on the THUMOS14 dataset, we set the NMS threshold $\rho$, confidence threshold $\psi$, IoU threshold $\eta_0$, $\eta_1$, $\eta_2$, and others as follows: $\rho$ = 0.45, $\psi$ = 0.1, $\eta_0$ =0.4, $\eta_1$ =0.4, $\eta_2$ =0.1, $\alpha$ = 0.5, $\beta$ =  n. For experiments on the ActivelyNet v1.2 and v1.3 datasets, we set the NMS threshold to 0.9, keeping the other parameters consistent with the THUMOS14 dataset. All experiments were conducted on a single Tesla V100.

The sensitivity analysis line chart for all hyperparameters is shown below:
\begin{figure}[H]
        \centering
        \tiny
        \resizebox{\columnwidth}{!}{
        \begin{tabular}{cc}
        \renewcommand{\thefigure}{4}
        \begin{minipage}[t]{2in}
        \includegraphics[width=2in]{fig/Hyperparametric Sensitivity Analysis (ρ).png}
        \caption{ NMS threshold $\rho$.}
        \end{minipage}
    \renewcommand{\thefigure}{5}
        \begin{minipage}[t]{2in}
        \includegraphics[width=2in]{fig/Hyperparametric Sensitivity Analysis (ψ).png}
        \caption{ confidence threshold $\psi$.}
        \end{minipage}
    \renewcommand{\thefigure}{6}
        \begin{minipage}[t]{2in}
        \includegraphics[width=2in]{fig/Hyperparametric Sensitivity Analysis (η0).png}
        \caption{ IoU threshold $\eta_0$.}
        \end{minipage}
        \renewcommand{\thefigure}{7}
        \begin{minipage}[t]{2in}
        \includegraphics[width=2in]{fig/Hyperparametric Sensitivity Analysis (η1).png}
        \caption{ IoU threshold $\eta_1$.}
        \end{minipage}
    \end{tabular}}
\end{figure}
\begin{figure}[H]
        \centering
        \tiny
        \resizebox{\columnwidth}{!}{
        \begin{tabular}{cc}
        
    \renewcommand{\thefigure}{8}
        \begin{minipage}[t]{2in}
        \includegraphics[width=2in]{fig/Hyperparametric Sensitivity Analysis (η2).png}
        \caption{ IoU threshold $\eta_2$.}
        \end{minipage}
    \renewcommand{\thefigure}{9}
        \begin{minipage}[t]{2in}
        \includegraphics[width=2in]{fig/Hyperparametric Sensitivity Analysis (α).png}
        \caption{ hyper-parameter $\alpha$.}
        \end{minipage}
    \renewcommand{\thefigure}{10}
        \begin{minipage}[t]{2in}
        \includegraphics[width=2in]{fig/Hyperparametric Sensitivity Analysis (β).png}
        \caption{ hyper-parameter $\beta$..}
        \end{minipage}
    \renewcommand{\thefigure}{ }
        \begin{minipage}[t]{2in}
        \includegraphics[width=2in]{fig/null.png}
    
        \end{minipage}
    \end{tabular}}
    
\end{figure}

\section{Inference Pipeline.}
\label{section4}
During the inference process, the WTAL Base Model is not utilized. We directly input clip-level video features into the student model within the teacher-student training framework. Outputs are obtained from its classification and regression heads. Subsequently, we subject this output to Non-Maximum Suppression (NMS) to eliminate overlapping predictions and obtain the final action segments.
\section{Additional Visualization Results.}
\label{section5}
We provide more qualitative results in Figure 5. As depicted below, when the AIC module is introduced, redundant instances converge to corrected instances. Similarly, when the MIC module is added, missing action instances are compensated for.
\newpage
\begin{figure}[h]
	\centering

        \renewcommand{\thefigure}{11}
	\includegraphics[width=18cm]{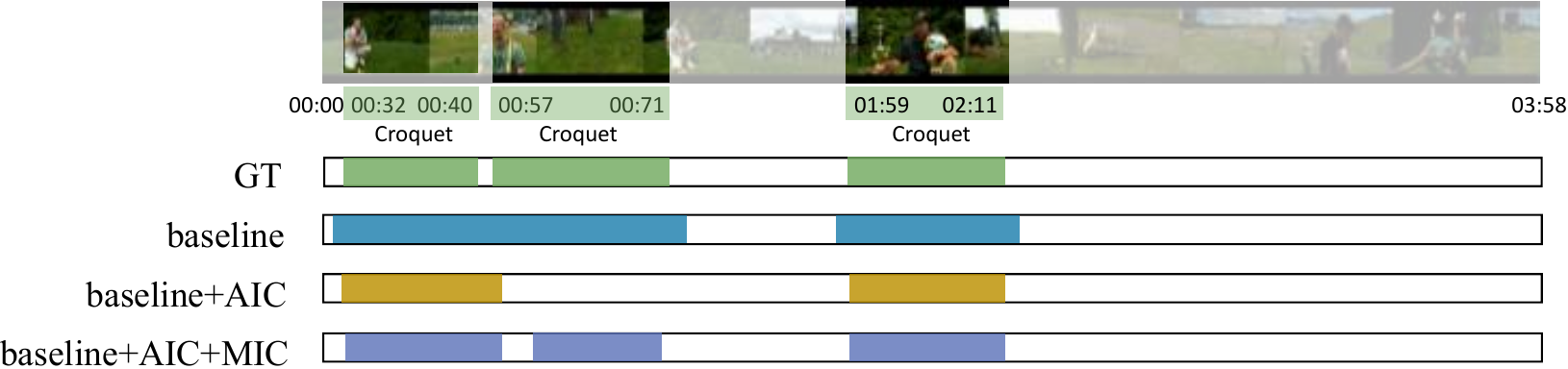}
	\caption{Visualization of ground-truth and predictions.}

	\label{Ablation-figure}
        \renewcommand{\thefigure}{12}
	\includegraphics[width=18cm]{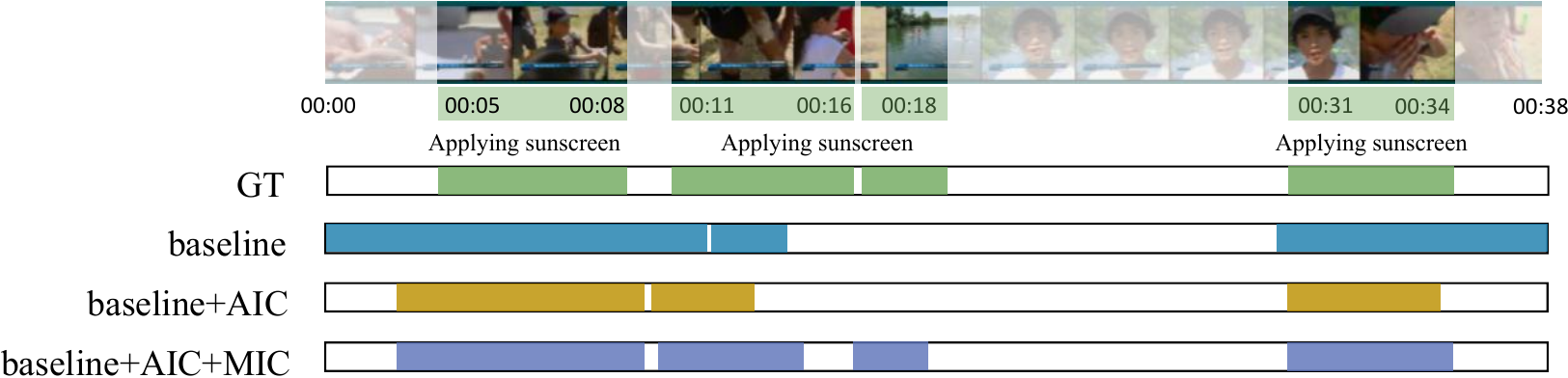}
	\caption{Visualization of ground-truth and predictions.}
	\label{Ablation-figure}

        \renewcommand{\thefigure}{13}
	\includegraphics[width=18cm]{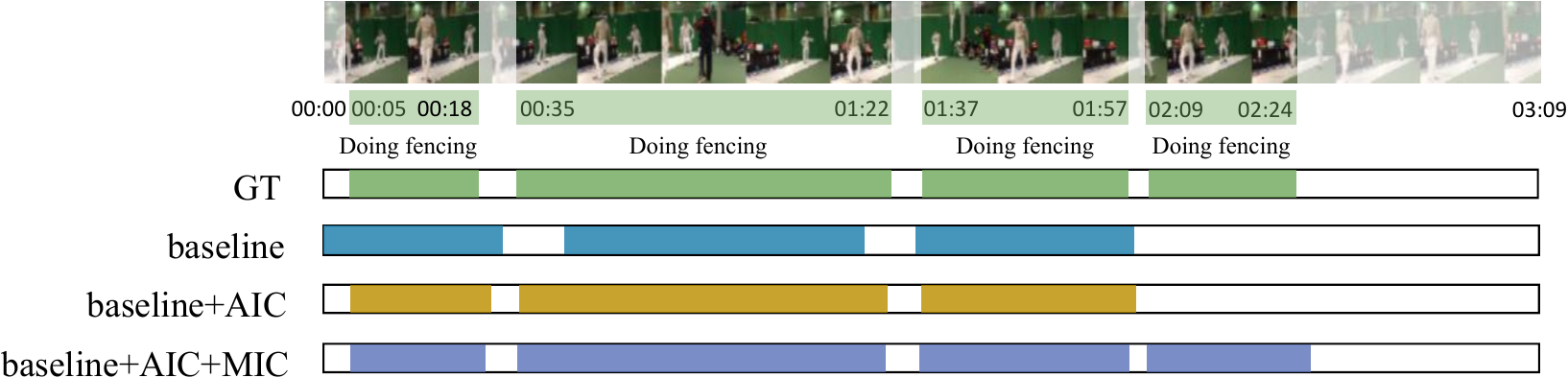}
	\caption{Visualization of ground-truth and predictions.}
\end{figure}
\newpage
\begin{figure}[h]
	\centering

        \renewcommand{\thefigure}{14}
	\includegraphics[width=18cm]{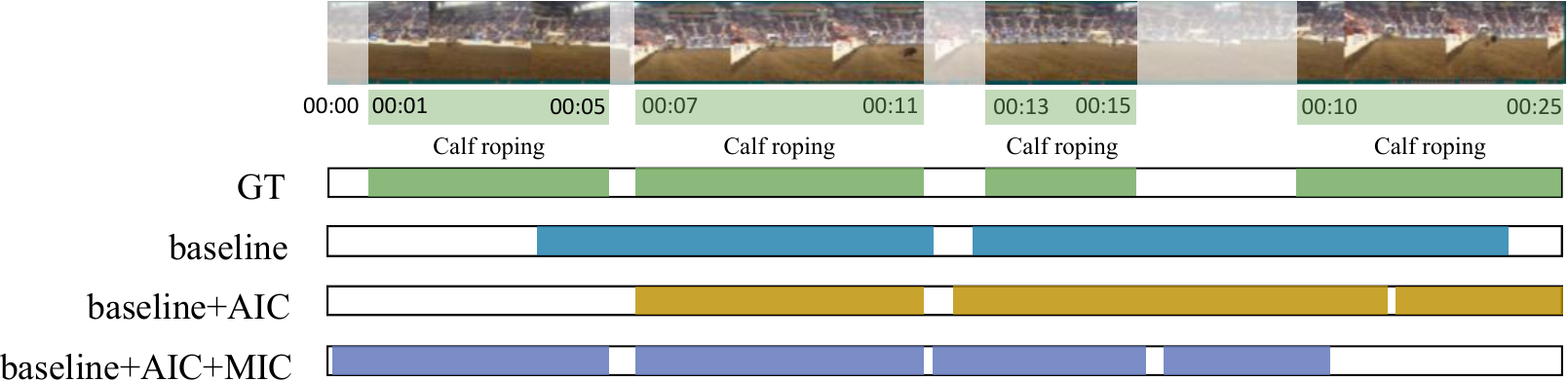}
	\caption{Visualization of ground-truth and predictions.}

	\label{Ablation-figure}
        \renewcommand{\thefigure}{15}
	\includegraphics[width=18cm]{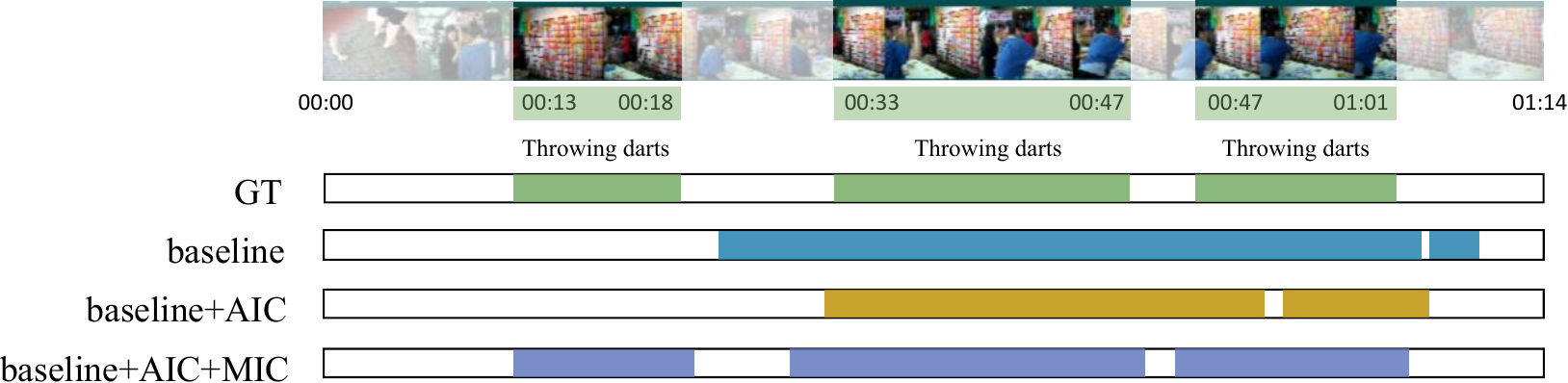}
	\caption{Visualization of ground-truth and predictions.}
	\label{Ablation-figure}

        \renewcommand{\thefigure}{16}
	\includegraphics[width=18cm]{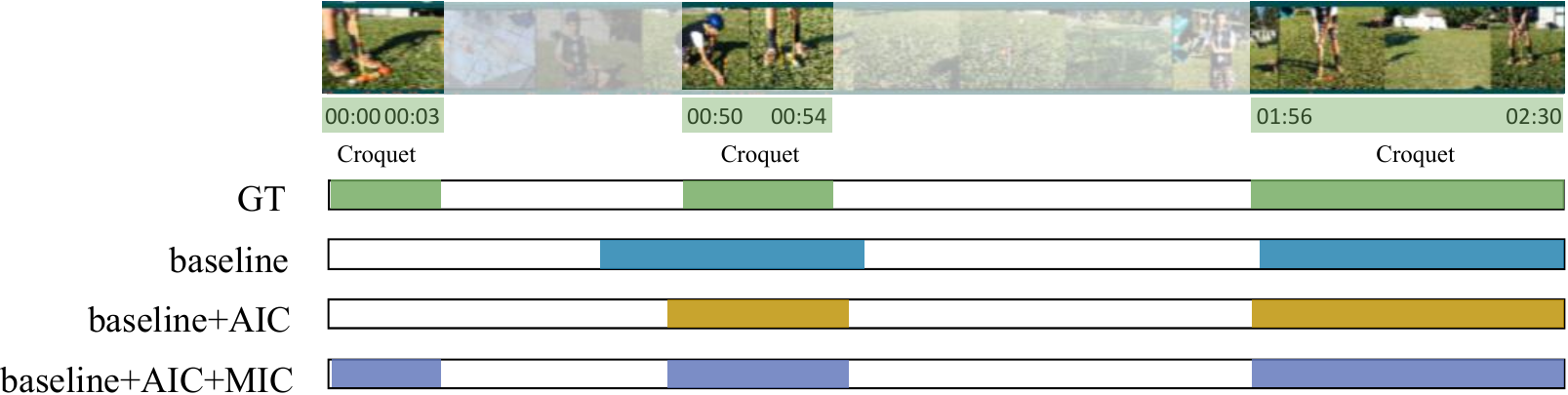}
	\caption{Visualization of ground-truth and predictions.}
\end{figure}
\newpage
\begin{figure}[h]
	\centering

        \renewcommand{\thefigure}{17}
	\includegraphics[width=18cm]{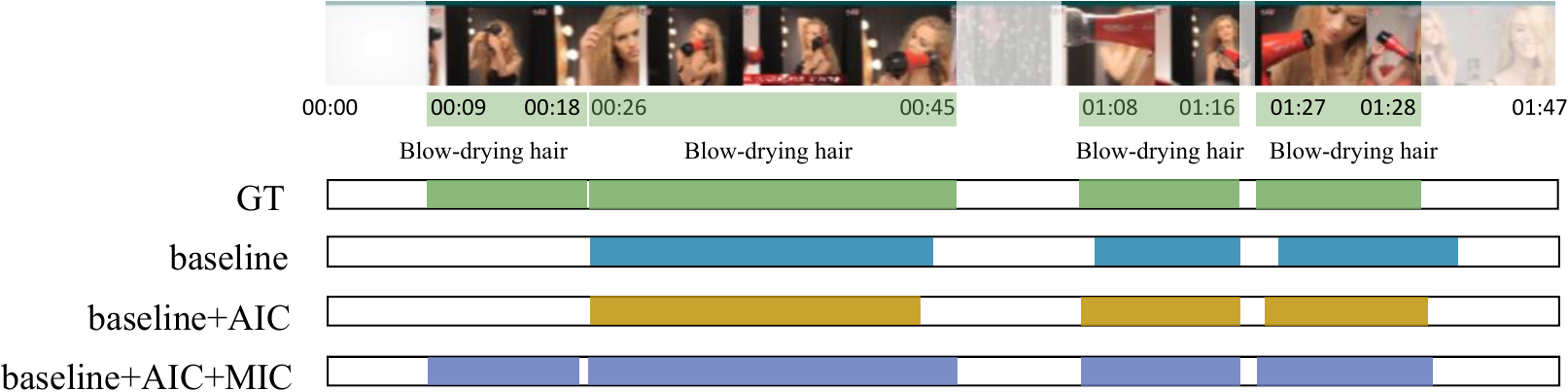}
	\caption{Visualization of ground-truth and predictions.}

	\label{Ablation-figure}
        \renewcommand{\thefigure}{18}
	\includegraphics[width=18cm]{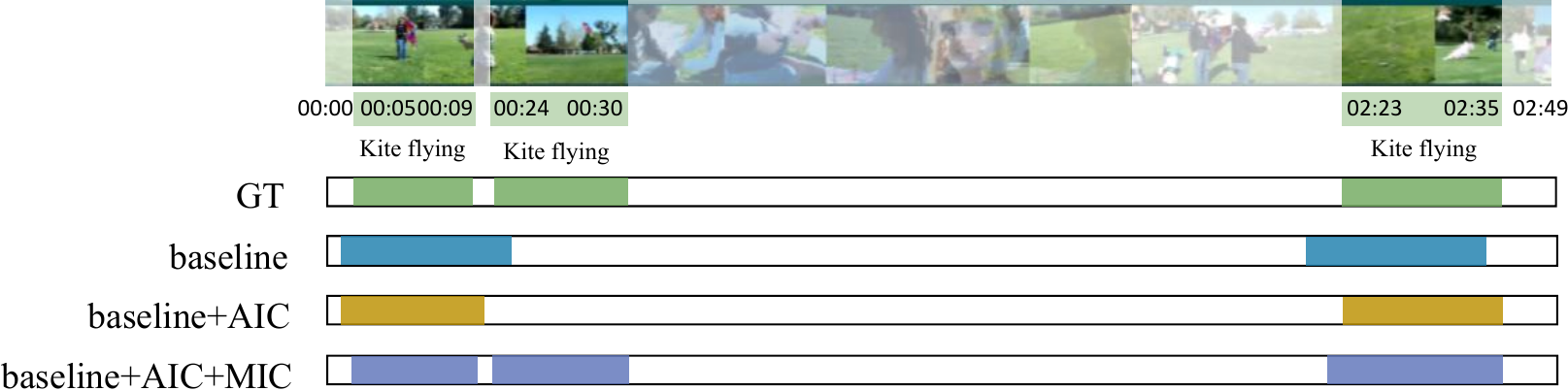}
	\caption{Visualization of ground-truth and predictions.}
	\label{Ablation-figure}
        \renewcommand{\thefigure}{19}
	\includegraphics[width=18cm]{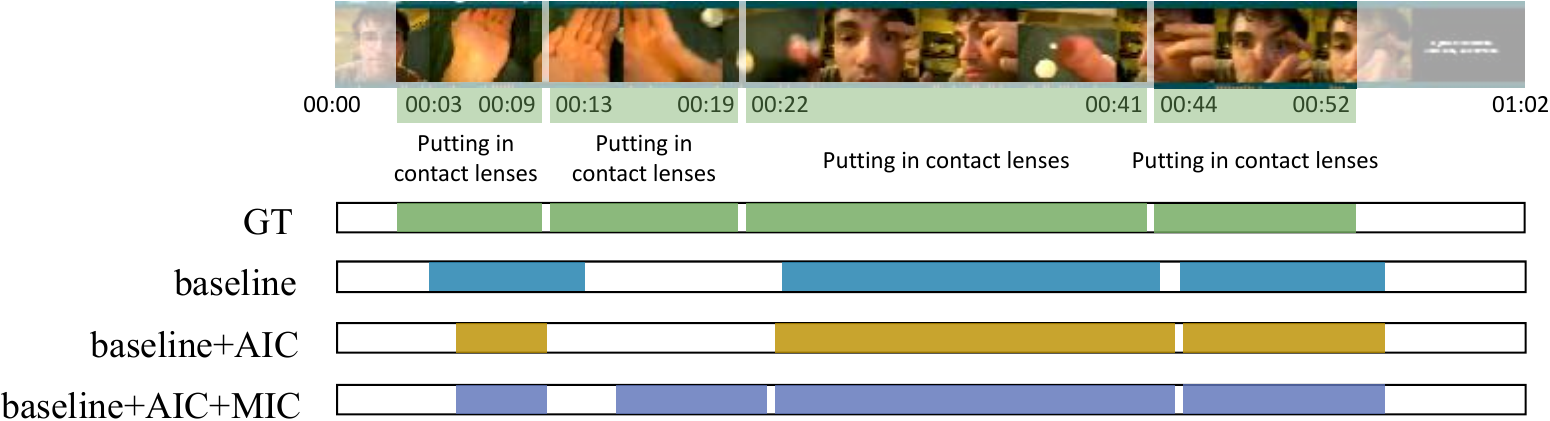}
	\caption{Visualization of ground-truth and predictions.}
\end{figure}
\newpage
\begin{figure}[h]
	\centering

        \renewcommand{\thefigure}{20}
	\includegraphics[width=18cm]{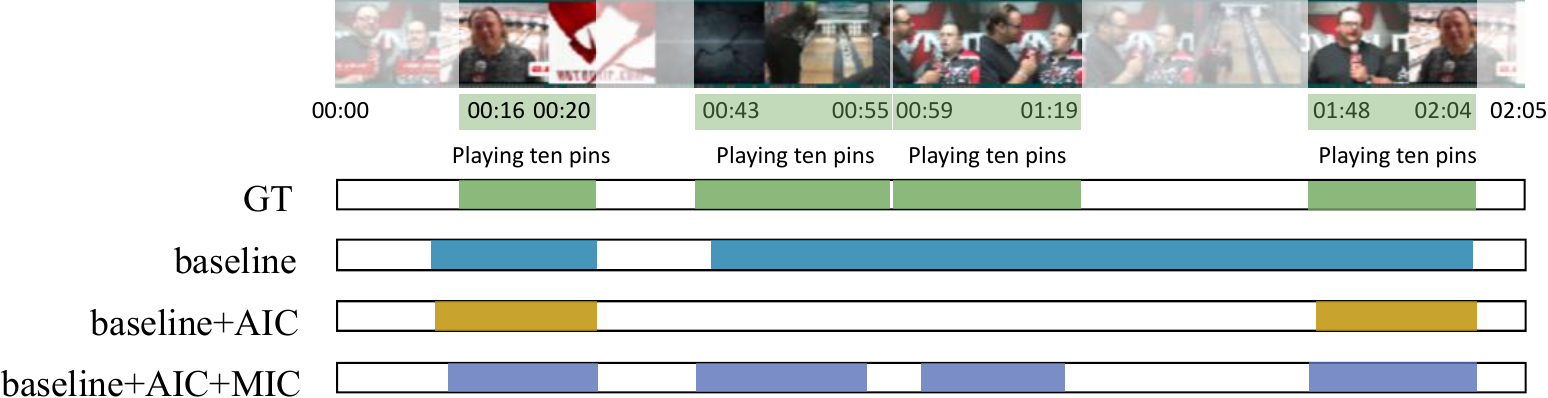}
	\caption{Visualization of ground-truth and predictions.}

	\label{Ablation-figure}
        \renewcommand{\thefigure}{21}
	\includegraphics[width=18cm]{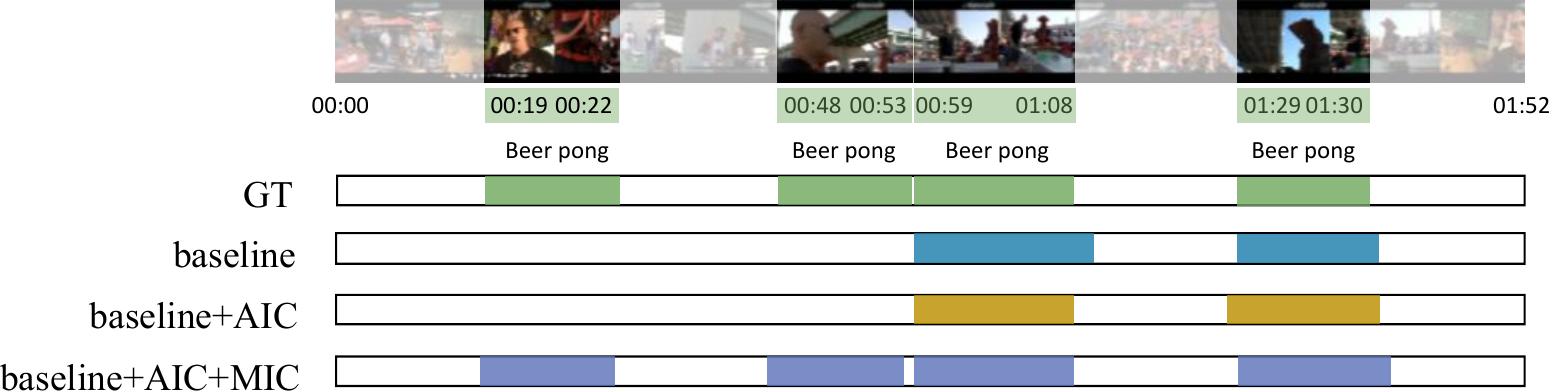}
	\caption{Visualization of ground-truth and predictions.}
	\label{Ablation-figure}
        \renewcommand{\thefigure}{22}
	\includegraphics[width=18cm]{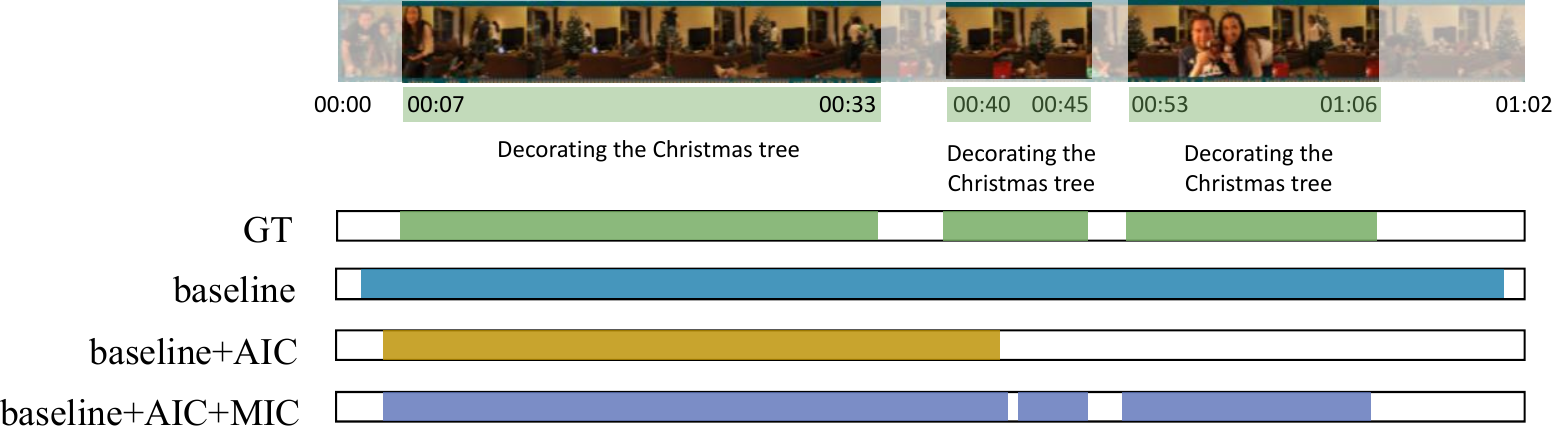}
	\caption{Visualization of ground-truth and predictions.}
\end{figure}

\newpage
\section{Broader Impacts.}
\label{section6}
As the predominant media format today, information is primarily conveyed through videos. The temporal action localization task involves identifying temporal boundaries and categorizing actions of interest in untrimmed videos. Unlike supervised learning, which relies on dense segment-level annotations, our proposed weakly-supervised temporal action localization model, NoCo, only requires video-level labels. This makes WTAL particularly valuable in real-world applications, such as popular video-sharing social-network services, where billions of videos are tagged solely with video-level user-generated labels. Additionally, WTAL finds broad applications in various fields, including event detection, video summarization, highlight generation, and video surveillance.